\documentclass[runningheads]{llncs}
\usepackage[T1]{fontenc}
\usepackage{graphicx}
\usepackage{booktabs}
\usepackage[misc]{ifsym}
\newcommand{\corr}{(\Letter)}

\usepackage{amsmath}

\usepackage{caption}
\usepackage{graphicx}
\usepackage{float}
\usepackage{subcaption}

\usepackage{algorithm}
\usepackage{algorithmic}

\usepackage{mwe}

\begin{document}

\title{GAP: Graph-Assisted Prompts for Dialogue-based Medication Recommendation}

\titlerunning{Graph-Assisted Prompts for Dialogue-based Medication Recommendation}

\author{Jialun Zhong\inst{1,2} \and Yanzeng Li\inst{1} \and Sen Hu\inst{2} \and Yang Zhang\inst{2} \and \\ Teng Xu\inst{2} \and Lei Zou\inst{1} \corr}


\authorrunning{Zhong et al.}

\institute{Wangxuan Institute of Computer Technology, Peking University, Beijing, China
\and
Ant Group \\ \email{zhongjl@stu.pku.edu.cn, zoulei@pku.edu.cn}}

\maketitle              

\begin{abstract}


Medication recommendations have become an important task in the healthcare domain, especially in measuring the accuracy and safety of medical dialogue systems (MDS). 
Different from the recommendation task based on electronic health records (EHRs), dialogue-based medication recommendations require research on the interaction details between patients and doctors,  which is crucial but may not exist in EHRs. 
Recent advancements in large language models (LLM) have extended the medical dialogue domain. These LLMs can interpret patients' intent and provide medical suggestions including medication recommendations, but some challenges are still worth attention.
During a multi-turn dialogue, LLMs may ignore the fine-grained medical information or connections across the dialogue turns, which is vital for providing accurate suggestions. Besides, LLMs may generate non-factual responses when there is a lack of domain-specific knowledge, which is more risky in the medical domain.
To address these challenges, we propose a \textbf{G}raph-\textbf{A}ssisted \textbf{P}rompts (\textbf{GAP}) framework for dialogue-based medication recommendation. 
It extracts medical concepts and corresponding states from dialogue to construct an explicitly patient-centric graph, which can describe the neglected but important information.
Further, combined with external medical knowledge graphs, GAP can generate abundant queries and prompts, thus retrieving information from multiple sources to reduce the non-factual responses.
We evaluate GAP on a dialogue-based medication recommendation dataset and further explore its potential in a more difficult scenario, dynamically diagnostic interviewing. Extensive experiments demonstrate its competitive performance when compared with strong baselines.


\keywords{Medication Recommendation \and Large Language Model \and Graph.}
\end{abstract}

\section{Introduction}
\begin{figure*}[ht]
    \centering
    \includegraphics[width=0.5\linewidth]{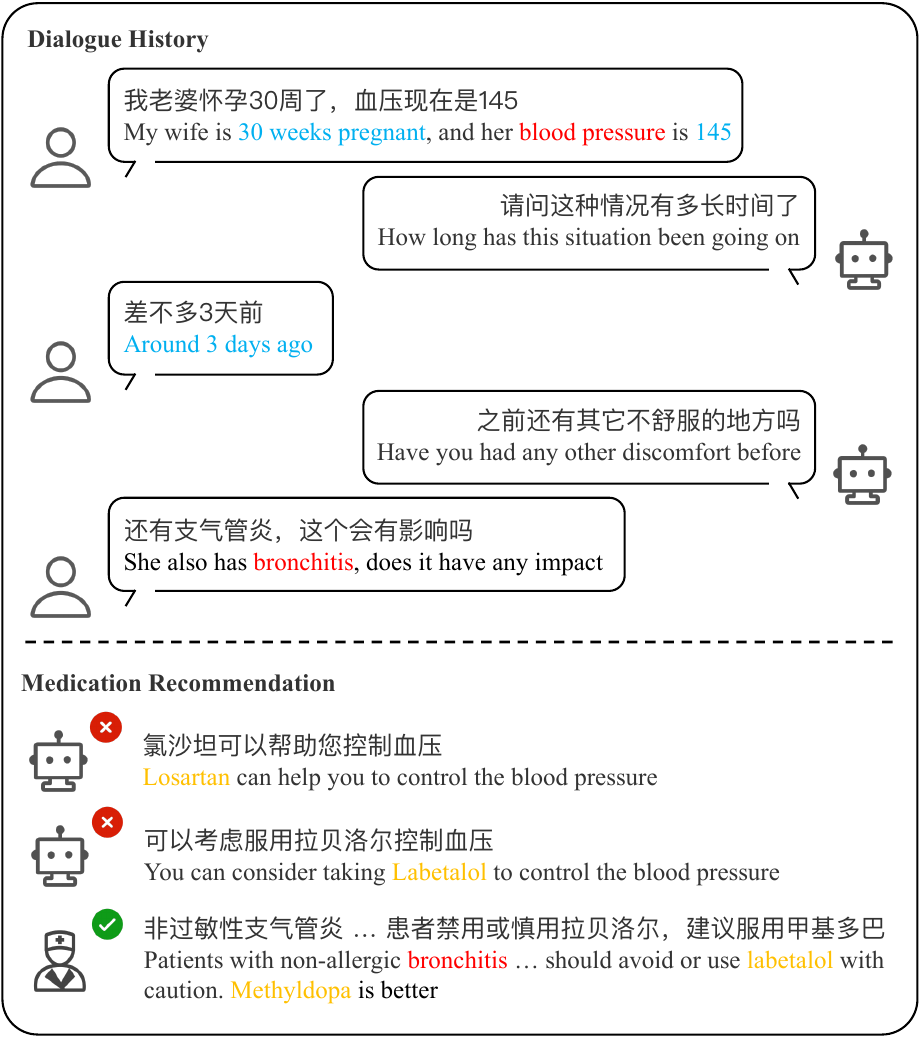}
    \caption{A medical consultation between a human patient and LLMs. Red, blue, and yellow represent diseases, states, and medication concepts. \textbf{Losartan} is restrained for \textbf{pregnancy}, and \textbf{Labetalol} must be used with caution when the patient suffers from \textbf{bronchitis}.}
    \label{figure:chat}
\end{figure*}



In response to the increasingly growing healthcare demands, medical dialogue systems (MDS)~\cite{DBLP:journals/corr/abs-2405-10630} have gained research interests due to their practical potential. 
An MDS can automatically collect and analyze patients' information in interactive manner and then provide corresponding medical services~\cite{DBLP:conf/emnlp/ZengYJYWZZZDZFZ20,DBLP:conf/coling/HeHOGCX022},
where medication recommendation~\cite{DBLP:conf/coling/HeHOGCX022} plays a pivotal role in the dialogue process. 
Only accurate and trustworthy recommendations can ensure patient well-being and safety \cite{DBLP:journals/corr/abs-2310-05694}, especially when the patient lacks knowledge in the medical domain.

Recently, the emergence of generative Large Language Models (LLMs)~\cite{ChatGPT} promotes the advance of academia and society, where medical domain~\cite{DBLP:journals/corr/abs-2310-05694} is no exception.
By pre-training on extensive medical text corpora, LLMs have accumulated plenty of medical knowledge and demonstrated the value in various medical tasks, including question answering~\cite{DBLP:journals/ijis/GuoCY22}, reports generation~\cite{DBLP:journals/corr/abs-2302-07257}, etc. 
Naturally, these abilities make LLMs well-adapted to the aforementioned services in MDS. Some recent studies~\cite{DBLP:journals/corr/abs-2304-01097,DBLP:journals/corr/abs-2306-09968} have explored LLMs in different dialogue-based applications.


However, there are still some challenges about directly utilizing LLMs for dialogue-based medication recommendations.
LLMs may struggle to accurately capture the nuanced medical information in the dialogue context, which is necessary for providing reliable recommendations. 
For example, the first robot doctor in Figure~\ref{figure:chat} neglects the fine-grained medical information ``30 weeks pregnant'' in the dialogue history~\cite{DBLP:conf/eacl/ZhuMBLPLZYT24} and recommends Losartan, which is contraindicated medication to pregnancy.
Additionally, without sufficient domain-specific knowledge, LLMs may generate responses containing inaccurate or unfounded claims, which can be hazardous in the context of recommendation guidance. 
For example, the second response in Figure~\ref{figure:chat} ignores the restriction on using Labetalol when the patient is suffering from bronchitis, resulting in the risk of medical malpractice.



To address the challenges of knowledge-intensive response generation in medical dialogue applications, some LLMs~\cite{DBLP:journals/corr/abs-2308-14346} 
have utilized knowledge-based instruction data in training stage.
It remains a struggle for LLMs to memorize the long-tail medical knowledge~\cite{DBLP:conf/icml/KandpalDRWR23} (e.g., drug interactions) that can be crucial for recommendations.
Other recent works have explored alternative approaches like In-Context Learning (ICL)~\cite{DBLP:conf/emnlp/DouJJZZT23} and Retrieve Augmented Generation (RAG)~\cite{DBLP:journals/corr/abs-2312-15883}, to integrate with extra information source outside LLMs. However, these methods may overlook the fine-grained medical information across the dialogue history, potentially leading to hallucinations~\cite{DBLP:journals/corr/abs-2401-01313}.

In this work, we propose GAP, a \textbf{G}raph-\textbf{A}ssisted \textbf{P}rompts framework for dialogue-based medication recommendations. It enhances the comprehension of dialogue history, meanwhile infusing LLMs with medical knowledge. GAP is a plug-and-play framework that is orthogonal with the aforementioned methods.
Specifically, GAP extracts medical mentions (e.g., diseases, medications, etc.) and their corresponding states from the dialogue flow.
Then GAP normalizes the mentions and maintains a patient-centric dialogue graph explicitly. The graph can be used to measure whether the LLMs comprehend and memorize the medical information in dialogue history. 
GAP further combines medical Knowledge Graphs (KGs) with the patient-centric graph to generate context-relevant queries. It then produces prompts for LLMs by executing the queries on multiple knowledge sources. This allows for the integration of specialized background medical knowledge in an RAG manner, therefore avoiding irresponsible responses.


Our main contributions are as follows:
\begin{itemize}
    \item We propose a novel LLM-based medication recommendation framework, GAP, which constructs patient-centric graphs for capturing and maintaining fine-grained medical information, including medical concepts and states.
    \item We introduce external knowledge sources into GAP by integrating medical-specific knowledge graphs and producing auxiliary and informative prompts based on the patient-centric graph, further improving the professionalism of medication recommendation.
    \item We conduct experiments on the medication recommendation task and further explore GAP's potential on dynamically diagnostic interviewing. The results indicate the effectiveness of our framework.
\end{itemize}


\section{Related Work}
\subsection{Medical Dialogue Systems}
Medical dialogue systems can automatically analyze patients' intents and provide medical-related assistance, which has achieved prominent results in different scenarios like 
consultations~\cite{DBLP:conf/emnlp/ZengYJYWZZZDZFZ20}, 
medical education~\cite{DBLP:journals/corr/abs-2404-13066}, and medication recommendation~\cite{DBLP:conf/coling/HeHOGCX022}. The elaborated pipeline MDSs~\cite{valizadeh-parde-2022-ai} contain extra components (e.g., dialogue state tracking, external knowledge augmentation) for natural language understanding than end-to-end MDSs. For example, HMIE~\cite{DBLP:conf/hpcc/ZhaoJZML22} extracts heterogeneous information from dialogue history to model dialogue topics and doctor actions. MedPIR~\cite{DBLP:conf/kdd/ZhaoLWHC0D022} designs a knowledge-aware dialogue graph encoder and focuses on pivotal information in the long history context. Besides, DFMed~\cite{DBLP:conf/acl/XuHCWL23} considers the transitions of medical entities and doctor actions, and generates the flow-guided response. Recently, the emergence of LLMs also drives the advance of MDS. Some studies focus on the medical training strategy, such as 
self-chat data collection~\cite{DBLP:conf/emnlp/XuGDM23} and reinforced learning from AI feedback~\cite{DBLP:conf/emnlp/ZhangCJYCCLWZXW23}. Other works utilize prompting technology like ICL~\cite{DBLP:conf/emnlp/DouJJZZT23} and hypothesis output~~\cite{DBLP:journals/corr/abs-2312-15883}, which pay more attention to enhance the inference ability of LLMs.


\subsection{Medication Recommendation}
Medication recommendation methods aim to make trustworthy medication prescriptions based on the patient situation. Previous works mainly focus on the recommendation task according to EHRs. For example, LEAP~\cite{DBLP:conf/kdd/ZhangCTSS17} decomposes the recommendation into a sequential decision-making process and determines the medications. DMNC~\cite{DBLP:conf/kdd/Le0V18} uses a dual memory neural computer to model asynchronous multi-view sequential information. GAMENet~\cite{DBLP:conf/aaai/ShangXMLS19} introduces a graph memory module and models longitudinal patient records to provide recommendations. DKINet~\cite{liu2023dkinet} proposes a knowledge injection module to integrate patient clinical manifestations and domain knowledge to describe health conditions. LEADER~\cite{DBLP:journals/corr/abs-2402-02803} fine-tunes a LLM-based medication recommendation model and then distills it to acquire a student model for high-efficiency inference.
Different from the aforementioned methods, DDN~\cite{DBLP:conf/coling/HeHOGCX022} makes the first attempt to dialogue-based medication recommendation, which combines the dialogue and a disease representation to make the prediction.

\subsection{Retrieval Augmented Generation}
Retrieval Augmented Generation (RAG)~\cite{DBLP:conf/nips/LewisPPPKGKLYR020} is feasible to enhance LLMs in knowledge-intensive tasks~\cite{DBLP:journals/corr/abs-2312-10997} to mitigate hallucination~\cite{DBLP:journals/corr/abs-2401-01313}. These methods retrieve external knowledge as context, which combines with the original query to formulate a coherent prompt, and finally leads to the response. The source of knowledge acquisition can be diverse, such as relevant examples~\cite{DBLP:conf/emnlp/DouJJZZT23}, web documents~\cite{DBLP:journals/corr/abs-2303-01229}, and knowledge graphs~\cite{DBLP:journals/corr/abs-2311-17330}. Some efforts attempt to improve the RAG system from different aspects. For example, LLM4CS~\cite{DBLP:conf/emnlp/MaoDMH0Q23} uses an LLM-based query rewriter to robustly represent the contextual intent in multi-turn dialogue. RECOMP~\cite{DBLP:journals/corr/abs-2310-04408} presents LM-based compressors to select relevant information from retrieved documents. Self-Ask~\cite{DBLP:conf/emnlp/PressZMSSL23} utilizes LLMs to actively state the questions and access an internet search engine. Compared with the aforementioned studies, GAP focuses on mining useful information and potential risks from graph data during pre-generation, thereby mitigating the harmfulness of LLMs.

\section{Methodology}


This section describes the details of our framework. GAP first extracts medical concept mentions and corresponding states from the origin dialogue to participate in the construction of a patient-centric graph. Then it generates three different types of prompts based on the graph and leverages them for target response generation. The overall architecture is shown in Figure \ref{figure:architecture}.

\subsection{Notations} \label{sec 3.1}

\paragraph{Knowledge Graph.} Let knowledge graph $\mathcal{G} = (\mathcal{E}, \mathcal{R}, \mathcal{F})$, where $\mathcal{E}$, $\mathcal{R}$, $\mathcal{F}$ denote the sets of entities, relations, and facts, respectively. A fact in $\mathcal{F}$ can be represented as $(e_{head}, r, e_{tail})$, where $e_{head}, e_{tail} \in \mathcal{E}$, and $r \in \mathcal{R}$. The triplet stands for a directed edge in $\mathcal{G}$ of type $r$ from node $e_{head}$ to node $e_{tail}$. Besides, $\mathcal{N}_\mathcal{G^\prime} = \{(r,e)|(e^\prime, r,e)\in \mathcal{F}, e^\prime\in \mathcal{G^\prime}\}$ is defined as the neighborhood of a subgraph $\mathcal{G^\prime} \subset \mathcal{G}$.

\paragraph{Medical Dialogue.} Let $\mathcal{H} = (p_1, d_1, ..., p_{i})$ be a multi-round medical dialogue between a patient and a doctor, where utterances $\mathcal{P} = (p_1, ..., p_{i})$ and $\mathcal{D} = (d_1, ..., d_{i-1})$ belong to patient and doctor respectively. An MDS aims to generate $d_i$ which contains diagnosis or recommendation.

\subsection{Medical Information Extraction} \label{sec 3.2}

Capturing key medical information from the dialogue history can play a crucial role in MDS~\cite{DBLP:conf/aaai/ShiHCSLH20}.
The extracted information can be used for relevant knowledge retrieval to enhance the response generation. Inspired by previous works \cite{DBLP:journals/corr/abs-2302-10205}, 
we adopt a two-stage method for medical information extraction. Without loss of generality, for patient's utterance $p_m$, we first identify all medical concepts $\mathcal{C}$ (e.g., disease, symptom) it contains by utilizing LLMs and instruction $\mathcal{I}_{NER}$ consists of a prompt and a few demonstrations:

\begin{equation}
\begin{split}
    \mathcal{C} = \mathtt{LLM}(\mathcal{I}_{NER}, p_m), 
\end{split}
\end{equation}

Given a concept $c$ in $p_m$. We then judge its states by formulating it as a slot-filling task. The slots are pre-defined based on the category of $c$. Considering that the state mentions can be scattered in multi-turn utterances due to the patients may lack medical knowledge and cannot express clearly \cite{DBLP:conf/emnlp/DouJJZZT23}, we extend the surrounding utterances of $p_m$ to the context and obtain the slot-value pairs $\mathcal{SV}$ of $c$:

\begin{equation}\label{equ:sv}
\begin{split}
    \mathcal{SV} &= \{(c,s,v)|c\in \mathcal{C}\} \\
                &= \mathtt{LLM}(\mathcal{I}_{SV}, p_{m-k:m+k}, d_{m-k:m+k}), 
\end{split}
\end{equation}

\begin{figure*}[t]
    \centering
    \includegraphics[width=\linewidth]{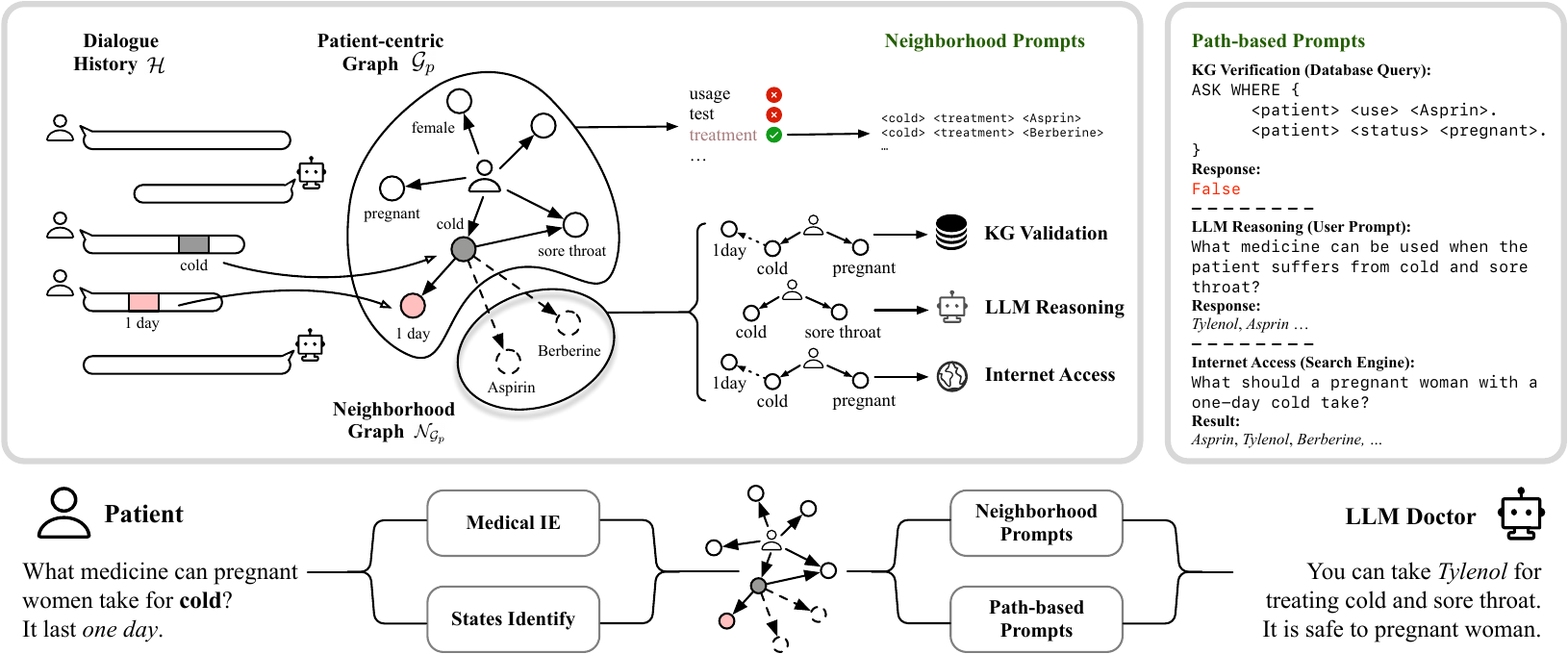}
    \caption{Overall diagram of GAP.}
    \label{figure:architecture}
\end{figure*}

\subsection{Graph Construction} \label{sec 3.3}

The extracted medical information can be used to build up a graph as the medical dialogue memory, which has been proven to benefit clinical downstream applications \cite{DBLP:conf/emnlp/GaoZWG023}. We call it a patient-centric graph (denoted as $\mathcal{G}_p$) as it contains explicit \textit{patient} nodes and associated medical concept nodes, designed to accommodate multiple patients probably mentioned in the whole dialogue. Thus, the graph is initialized with a \textit{patient} node. For medical concept $c\in\mathcal{C}$ in utterance $p_m$, a node $c$ and an edge from the patient node to $c$ are added to the graph. The relation depends on the medical type of $c$. Analogously, the slot-value pair $(c,s,v) \in \mathcal{SV}$ is converted to a directed edge from $c$ to $v$ with type $s$. The expansion and update of the patient-centric graph are under the dialogue flow.

Considering that one medical concept can have mentions with subtle differences, we use simple linking methods (e.g., edit distance, synonym list) to transform the various mentions into a normalized name in available KGs. Therefore, the related KG entities and edges can naturally be introduced as a neighborhood graph $\mathcal{N}_{\mathcal{G}_p}$. In the top-left part of Figure \ref{figure:architecture}, the dotted line denotes KG entities ``Aspirin'' and ``Berberine'' which are the drug treatments (omitted in Figure) of dialogue entity ``cold''. The graph now contains key information from dialogue and KGs, prompting LLMs to generate responses.

\subsection{Prompt Generation} \label{sec 3.4}

In this section, we introduce the prompts excavated from the patient-centric graph. Reasonable prompts should possess two qualities: summarizing key information from dialogue history and containing essential knowledge for response generation. To this end, we design neighborhood prompts and path-based prompts.

\subsubsection{Neighborhood Prompts}

The neighborhood prompts are used to retrieve knowledge from $\mathcal{N}_{\mathcal{G}_p}$. However, some studies \cite{DBLP:journals/corr/abs-2310-04408,DBLP:journals/corr/abs-2310-01558} demonstrate that simply using all the facts in $\mathcal{N}_{\mathcal{G}_p}$ as LLM's context may lead to computational costs and irrelevant information misuse. Filtering the retrieved knowledge is essential.

\begin{equation}
\begin{split}
    r_m = \mathtt{Top}1(\{r|(r_m,e^\prime)\in \mathcal{N}_{\mathcal{G}_p}, \mathcal{H} \}) \\
    \mathcal{NP} = \mathtt{Top}k_1(\{(e,r_m,e^\prime)| (r_m,e^\prime)\in \mathcal{N}_{\mathcal{G}_p}\}) 
\end{split}
\end{equation}

Inspired by previous pipeline MDS architectures, we first identify the knowledge type needed for current response generation, and then select the facts from $\mathcal{N}_{\mathcal{G}_p}$ with corresponding relation to get neighborhood prompts $\mathcal{NP}$. We leverage LLMs to achieve these operations.

\subsubsection{Path-based Prompts}

Different from neighborhood prompts, path-based prompts are designed to mine compositional and patient-specific path knowledge from the graph. The increase in the graph size results in a significant growth in the number of paths, hence we address this issue by reducing it to a path matching paradigm. Formally, we engage medical professionals to pre-define the prominent medical schema $\mathcal{S}$, and extract corresponding paths from each graph to get $\mathcal{PS}$:

\begin{algorithm}[t]
\caption{Methodology}\label{alg:methodology}
\algsetup{linenosize=\small}
  \small
\begin{algorithmic}[1]
\REQUIRE Knowledge Graph $\mathcal{G} = (\mathcal{E}, \mathcal{R}, \mathcal{F})$, \\Medical Dialogue History $\mathcal{H} = (p_1, d_1, ..., p_{m-1})$
\ENSURE Generated Doctor Response $d_m$

\STATE \textbf{Step 1: Medical Information Extraction}
\FOR{each patient's utterance $p_m$ in $\mathcal{P}$}
    \STATE Extract medical concepts $\mathcal{C}$ using LLM with instruction $\mathcal{I}_{NER}$:
    \STATE $\mathcal{C} \gets \mathtt{LLM}(\mathcal{I}_{NER}, p_m)$
    
    \FOR{each concept $c$ in $\mathcal{C}$}
        \STATE Extend context to surrounding utterances $p_{m-k:m+k}$ and $d_{m-k:m+k}$
        \STATE Extract slot-value pairs $\mathcal{SV}$ for $c$ using LLM with instruction $\mathcal{I}_{SV}$:
        \STATE $\mathcal{SV} \gets \mathtt{LLM}(\mathcal{I}_{SV}, p_{m-k:m+k}, d_{m-k:m+k})$
    \ENDFOR
\ENDFOR

\STATE \textbf{Step 2: Patient-Centric Graph Construction}
\STATE $\mathcal{G}_{patient} \gets \mathcal{C} \bigoplus \mathcal{SV}$

\STATE \textbf{Step 3: Prompt Generation}
\STATE $\mathcal{NP} \gets $ Retrieve Neighborhood from $\mathcal{N}_{\mathcal{G}_p}$ based on $\mathcal{H}$
\STATE $\mathcal{PS} \gets $ Construct Path Query from $\mathcal{G}_p$ and $\mathcal{N}_{\mathcal{G}_p}$
\STATE $\mathcal{PP} \gets $ Retrieve Information from LLM, Internet, KG with query $\mathcal{PS}$ 

\STATE \textbf{Step 4: Response Generation}
\STATE $d_m \gets \mathtt{LLM}(\mathcal{H}, \mathcal{G}_p, \mathcal{NP}, \mathcal{PP})$
\RETURN $d_m$
\end{algorithmic}
\end{algorithm}

\begin{equation}
\begin{split}
    \mathcal{PS} = \mathtt{Top}k_2(\mathtt{Match}(\mathcal{S}, \mathcal{G}_p, \mathcal{N}_{\mathcal{G}_p})) 
\end{split}
\end{equation}

These paths can be viewed as queries to retrieve knowledge from different knowledge sources $\mathcal{KS}$ such as KGs, LLMs, and the Internet, which we introduce as follows.

\begin{equation}
\begin{split}
    \mathcal{PP} = \mathtt{Retrieve}(\mathcal{KS}, \mathcal{PS})
\end{split}
\end{equation}

\paragraph{KG Verification}

The entity attributes stored in medical KGs can be leveraged for validation. For instance, the attributes such as ``contraindications'', ``interactions'', and ``usage'' of \textit{drugs} exclude unsuitable medication candidates for recommendation when associated information (e.g., pregnant) is documented in the patient-centric graph. Consequently, the KG validation can aid in the elimination of risk for subsequent reasoning. The results of verification process are expressed as natural language prompts using pre-defined templates.

\paragraph{LLM Reasoning}

Another natural approach is to leverage the powerful ability of LLM for self-prompting. Some previous works have focused on generating auxiliary queries as intermediate reasoning steps, such as sub-questions \cite{DBLP:conf/emnlp/PressZMSSL23} and abstract questions \cite{DBLP:journals/corr/abs-2310-06117} to improve the accuracy. Our work is in analogy to these methods while using the path extracted from the patient-centric graph as prior information to improve the generated queries relevance to current intent. 


\paragraph{Internet Access}

Note that the knowledge in KGs can be missing or outdated, which can hinder the complex reasoning of LLMs. Besides, KGs hardly contain comprehensive consultation and diagnostics examples, particularly in scenarios involving multiple medical conditions (e.g., treatment for hypertension in pregnant women with bronchitis), which can lead to hallucinations when LLMs lack relevant knowledge.
Therefore, we integrate Internet search access into our framework to further improve the recommendation. For paths in $\mathcal{PS}$, we rewrite them into search queries in natural language by LLMs, and obtain the results returned by the Internet to construct prompts.


\subsection{Inference} \label{sec 3.5}

The LLMs take as input a combination of dialogue history $\mathcal{H}$, patient-centric graph $\mathcal{G}_p$, neighborhood prompts $\mathcal{NP}$, and path-based prompts $\mathcal{PP}$ to generate response $d_m$ about medication prediction of the current turn. The symbolic facts in $\mathcal{G}_p$ are represented as triples, so we linearize them into a textual string by simply concatenating the subject, predicate, and object of the triples.

\begin{equation}
\begin{split}
    d_m = \mathtt{LLM}(\mathcal{H}, \mathcal{G}_p, \mathcal{NP}, \mathcal{PP})
\end{split}
\end{equation}

The detailed process is illustrated in Algorithm~\ref{alg:methodology}.

\section{Experiments}

\subsection{Experiment Setup.}

\paragraph{Datasets.} We conduct the experiments on two medical tasks. (1) DialMed~\cite{DBLP:conf/coling/HeHOGCX022} is a dataset for dialogue-based medication recommendation which we mainly focus on. The dataset is about three departments, respiratory, gastroenterology, and dermatology, which is constructed from Chunyu-Doctor\footnote{https://www.chunyuyisheng.com/} and DXY Drugs Database\footnote{http://drugs.dxy.cn/} for medical conversations and normalized medication targets. (2) LLM as Patients~\cite{DBLP:journals/corr/abs-2404-13066} is a novel method to evaluate diagnostic interviewing abilities of MDS. An MDS engages in multi-turn dialogue with an LLM-based simulated patient. Each dialogue is endowed with a pre-defined patient condition. The entire dialogue histories are evaluated in unified metrics to make sure fair comparison of different MDS.

\paragraph{Evaluation metrics.} We evaluate the medication recommendation accuracy on DialMed by using Jaccard and F1 metrics. 

\begin{equation}
\begin{split}
    Jaccard = \frac{1}{|D|}\sum_{k=1}^{|D|} \frac{|Y^{(k)} \cap \hat{Y}^{(k)}|}{|Y^{(k)} \cup \hat{Y}^{(k)}|},
\end{split}
\end{equation}

\begin{equation}
\begin{split}
    F1 = \frac{1}{|D|}\sum_{k=1}^{|D|} \sum_{k=1}^{|D|} \frac{2 \cdot P^{(k)} \cdot R^{(k)}}{P^{(k)}+R^{(k)}},
\end{split}
\end{equation}
where $|D|$ is the amount of test set, $Y^{(k)}$ and $\hat{Y}^{(k)}$ represent the ground truth and predicted medication set of the $k$-th dialogue. $P^{(k)}$ and $R^{(k)}$ represent the Precision and Recall rate of the $k$-th dialogue.

For the diagnostic task, we follow the setting of~\cite{DBLP:journals/corr/abs-2404-13066} by adopting the weighted coverage rate of \textit{Aspects} and \textit{Information} in the examination checklist to obtain the score of diagnostic process. The definition denotes as follows:

\begin{equation}
\begin{split}
    score = 0.3 * \textit{Aspects} + 0.7 * \textit{Information},
\end{split}
\end{equation}
where \textit{Aspects} represents the essential inquiries from MDS (e.g., whether asked symptom, allergy history), while \textit{Information} represents the key items elicited from LLM-based patient (e.g., whether obtained temperature, blood pressure from inquiries). The evaluation is based on actual simulated patient examination, and the scores come from the ensemble of 5 different LLM evaluators.

\paragraph{Baselines.} For the medication recommendation task, we compare our framework with two types of methods: (1) supervised methods (pre-trained language models), including HiTANet~\cite{DBLP:conf/kdd/LuoYXM20}, LSAN~\cite{DBLP:conf/cikm/YeLXM20}, DialogXL~\cite{DBLP:conf/aaai/ShenCQX21}, and DDN~\cite{DBLP:conf/coling/HeHOGCX022}. (2) unsupervised methods (LLM-based), including I/O prompt, Chain-of-Thoughts (CoT) prompt~\cite{DBLP:conf/nips/Wei0SBIXCLZ22}, and KG-RAG~\cite{DBLP:journals/corr/abs-2311-17330}. We use 
DeepSeek-V3~\cite{deepseekai2025deepseekv3technicalreport}
as the foundation LLM.

For the diagnostic task, we compare our framework with: (1) general LLMs, including ChatGPT-3.5-turbo~\cite{ChatGPT}, ERNIE-4-Bot~\cite{ERNIE}, 
and Mixtral-8x7B~\cite{DBLP:journals/corr/abs-2401-04088}. (2) medical-specific LLMs, including BianQue-2-6B~\cite{DBLP:journals/corr/abs-2310-15896}, and DISC-MedLLM-13B~\cite{DBLP:journals/corr/abs-2308-14346}. (3) human performance with varying degrees of medical background knowledge.



\begin{table*}[t]
\caption{Medication recommendation Jaccard and F1 scores (\%) for three departments on DialMed. All results are multiplied by 100 and averaged over five runs. Underline indicates the best performance of supervised methods.}
\renewcommand{\arraystretch}{1.35}
\resizebox{\linewidth}{!}{
\begin{tabular}{lllllllll}
\hline \hline
             & \multicolumn{2}{c}{All Data}                         & \multicolumn{2}{c}{Respiratory}                      & \multicolumn{2}{c}{Gastroenterology}                 & \multicolumn{2}{c}{Dermatology}                      \\ \hline
Model        & \multicolumn{1}{c}{Jaccard} & \multicolumn{1}{c}{F1} & \multicolumn{1}{c}{Jaccard} & \multicolumn{1}{c}{F1} & \multicolumn{1}{c}{Jaccard} & \multicolumn{1}{c}{F1} & \multicolumn{1}{c}{Jaccard} & \multicolumn{1}{c}{F1} \\ \hline
\multicolumn{9}{c}{Statistics}                                                                                                                                                                                                           \\ \hline
TF-IDF       & 21.25±0.41 & 35.05±0.56 & 16.06±0.44 & 27.68±0.66 & 23.85±0.40 & 38.52±0.52 & 28.84±0.14 & 44.77±0.17 \\ \hline
\multicolumn{9}{c}{Supervised Methods}                                                                                                                                                                                                   \\ \hline
HiTANet      & 30.75±0.69 & 44.57±0.83 & 22.01±1.04 & 33.62±1.44 & 33.95±1.26 & 48.39±1.26 & 39.17±1.93 & 53.41±2.21 \\
LSAN         & 34.33±0.58 & 46.14±0.45 & 26.11±1.06 & 38.89±1.01 & 39.28±0.22 & 52.49±0.62 & 50.29±1.24 & 57.90±1.09 \\
DialogXL     & 36.27±0.34 & 53.23±0.40 & 27.12±0.24 & 42.67±0.36 & 40.91±0.14 & 58.06±0.15 & 48.68±0.81 & 65.48±0.66 \\
DDN          & \underline{42.62±0.35} & \underline{59.77±0.34} & \underline{32.26±1.25} & \underline{48.77±1.43} & \underline{44.86±0.54} & \underline{61.93±0.52} & \underline{56.99±0.53}  & \underline{72.60±0.43}     \\ \hline
\multicolumn{9}{c}{LLM w/o External Knowledge}  \\ \hline
I/O          & 14.49±0.96 & 19.65±1.24 & 10.86±0.92 & 15.34±1.30 & 21.14±2.87 & 28.96±3.31 & 14.50±0.82 & 18.74±0.92 \\
CoT          & 24.05±0.11 & 28.86±0.13 & 20.24±0.28 & 26.57±0.32 & 28.66±0.46 & 34.93±0.52 & 21.23±0.28 & 26.19±0.44 \\
few-shot CoT & 28.55±0.15 & 36.19±0.23 & 22.06±0.16 & 29.28±0.16 & 37.99±0.89 & 47.85±0.11 & 29.28±0.11 & 35.11±0.24 \\ \hline
\multicolumn{9}{c}{LLM w/ External Knowledge}  \\ \hline
KG-RAG       & 30.15±0.63  & 42.44±0.96  & 27.48±0.19  & 38.83±0.55  & 42.01±0.87  & 52.45±1.21  & 34.60±1.05  & 44.93±1.40  \\
GAP (ours)   & \textbf{39.28±0.77}  & \textbf{54.27±1.13}  & \textbf{29.33±0.42}  & \textbf{43.38±0.45}  & \textbf{45.44±2.33}  & \textbf{57.17±3.00}  & \textbf{51.54±1.15}  & \textbf{68.30±1.89}  \\ \hline \hline
\end{tabular}
}

\label{table:medicine_predict}
\end{table*}

\subsection{Implementation details}

\paragraph{Baseline details.} For unsupervised methods in medication recommendations, the entire candidate medication list is provided to LLM. The target of the I/O prompt is to generate direct recommendation, while the CoT prompt aims to generate the patients' disease and corresponding medication sequentially. We construct one demonstration for each medical department in few-shot setting. For KG-RAG, we re-implement their design by retrieving relevant knowledge triples of medical concepts mentioned in dialogue, and adapt the LLM's output to the recommendation task. Other baseline results we present are reported in previous works~\cite{DBLP:conf/coling/HeHOGCX022,DBLP:journals/corr/abs-2404-13066}.

\begin{table}[]\small
\caption{Diagnostic interviewing results with LLM-based patients and evaluators.}
\centering
\renewcommand{\arraystretch}{1.1}
\begin{tabular}{lcc}
\hline\hline
Model                          & \multicolumn{1}{l}{Information Density} & \multicolumn{1}{l}{Score} \\ \hline
Mixtral-8x7B                   & 0.11                                    & 0.33                              \\
BianQue-2-6B                   & 0.14                                    & 0.25                              \\
DISC-MedLLM-13B                & 0.15                                    & 0.43                              \\
ERNIE-4-Bot                    & 0.13                                    & 0.37                              \\
ChatGPT-3.5-turbo             & 0.15                                    & 0.51                              \\
GAP                            & 0.19                                    & 0.63                              \\ \hline
Human (Non-medical background) & 0.15                                    & 0.45                              \\
Human (Expert)                 & 0.27                                    & 0.78                              \\ \hline\hline
\end{tabular}
\label{table:vsp_test}
\end{table}

\begin{table}[h]\small
\caption{Ablation study on DialMed. NP: Neighborhood Prompts. PP: Path-based Prompts.}
\centering
\renewcommand{\arraystretch}{1.1}
\begin{tabular}{lcc}
\hline\hline
Method                                              & Jaccard  & F1  \\ \hline
GAP                                                 & 39.28  & 54.27  \\ \hline
GAP w/o NP                          & 32.03  & 43.89  \\
GAP w/o PP                          & 33.66  & 48.90  \\
GAP w/o NP and PP                   & 26.71  & 36.24  \\ \hline\hline
\end{tabular}
\label{table:ablation}
\end{table}

\paragraph{GAP details.} We combine CMeKG (Chinese Medical Knowledge Graph)\footnote{https://cmekg.pcl.ac.cn/} and Disease-KB\footnote{https://github.com/nuolade/disease-kb} as the background knowledge base for the following experiments. Besides, some knowledge\footnote{https://www.yixue.com/} (medication, symptom, etc.) with regard to recommendation are integrated as supplements. DeepSeek-V3~\cite{deepseekai2025deepseekv3technicalreport}
is utilized for medical information extraction and prompt generation. 
The context sliding window size of extraction is set to 4 (i.e., $k = 1$ in Equation \ref{equ:sv}), and set to $\infty$ in diagnostic task and the recommendation task, respectively.
When using LLM in each stage, we construct 3 corresponding demonstrations for In-Context Learning. The maximum number of neighborhood prompts and path-based prompts is set to 3, and the temperature of LLM is set to 0.2.
For recommendation task, we extract the medication in the generated response and apply the evaluation metrics.

\subsection{Results and Discussion}

\paragraph{Medication recommendation Result.} Table \ref{table:medicine_predict} reports the performance on the test set of DialMed. As can be seen, GAP outperforms other LLM-based baselines on all department cases, and approaches the performance of supervised methods, DialogXL and LSAN.
We also observe that simply prompting LLM for medication recommendation doesn't perform well. The performance of few-shot CoT prompting is much lower than TF-IDF, which means LLMs face the challenge of comprehending the complex medical conditions within the dialogue history, especially when limited demonstrations are accessed. 
The improvement from KG-RAG and GAP indicates the importance of external knowledge.
Besides, considering that candidate medications with similar effects may impede accurate recommendation, which potentially contributing to the notable performance differences observed across the departments.

\begin{figure*}
    \centering
    \includegraphics[width=\linewidth]{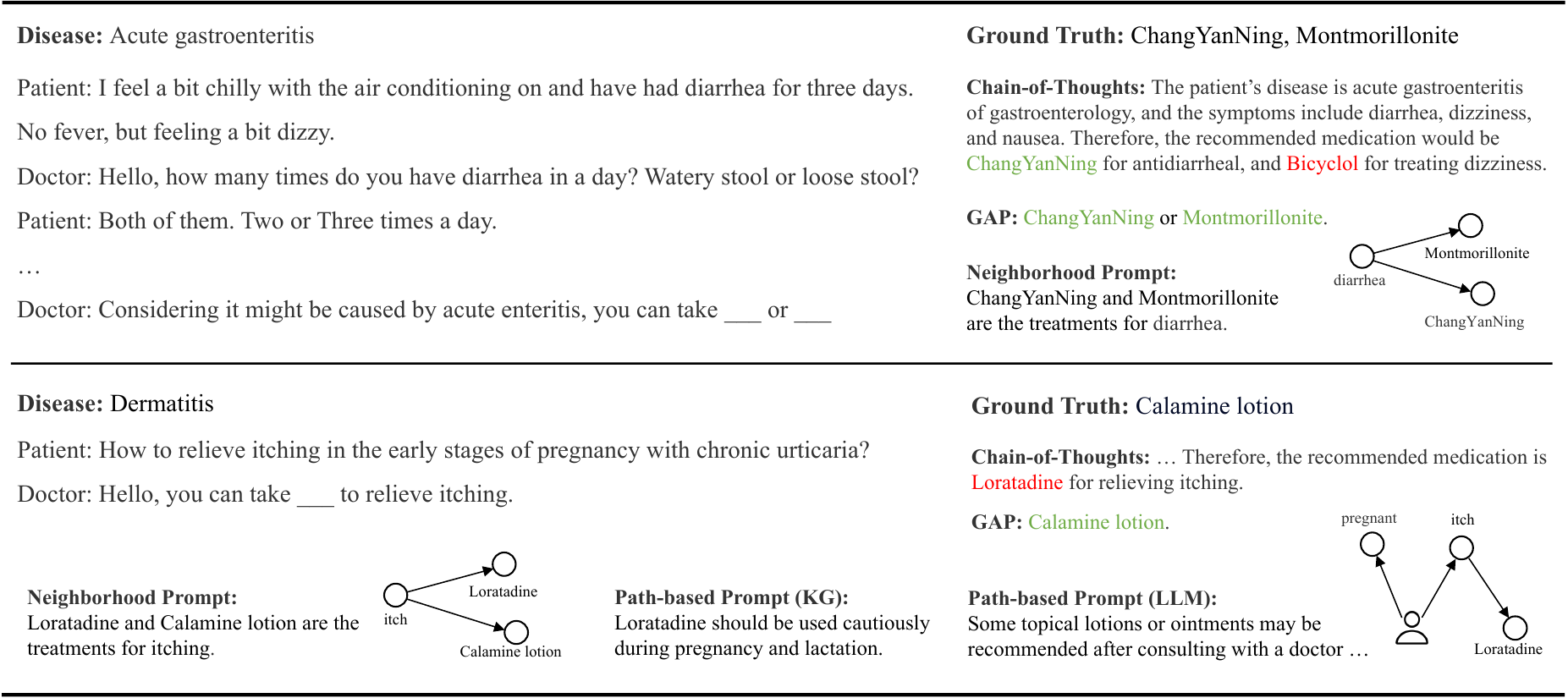}
    \caption{Medication recommendation cases on DialMed. GAP is compared with the Chain-of-Thoughts prompting strategy. Green and red indicate accurate and inaccurate recommendation respectively. The neighborhood prompts and path-based prompts are key elements that assist the recommendation.}
    \label{figure:cases}
\end{figure*}

\paragraph{Diagnostic interviewing Result.} Table \ref{table:vsp_test} shows the results of diagnostic interviewing evaluation. It can be observed that LLM's dynamic diagnostic ability in multi-turn scenario is weak relative to medical experts. The neighborhood prompts and path-based prompts designed in GAP can introduce prominent information to assist response generation based on the constructed graph, which can contributes to the performance gains

\begin{figure}
    \centering
    \includegraphics[width=0.7\linewidth]{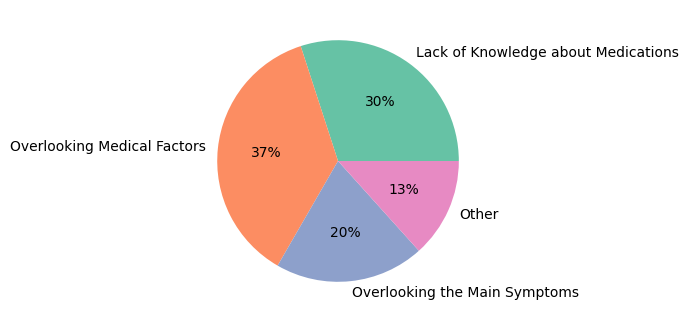}
    \caption{Error distribution on the samples of DialMed.}
    \label{figure:error}
\end{figure}


\paragraph{Ablation study.} In this paragraph, we study the effect of neighborhood prompts and path-based prompts. The results on DialMed are reported in Table \ref{table:ablation}. The variant of GAP without two kinds of prompts (only a patient-centric graph used for recommendation) outperforms the baselines of LLMs with no external knowledge, which indicates that the constructed graph can help LLMs comprehend the fine-grained medical concepts mentioned in the dialogue. 
The removal of neighborhood prompts can lead to a 19.13\% drop on F1, which denotes that external knowledge is a key component to ensure accurate recommendations.
Besides, GAP outperforms the variant without path-based prompts, which means an improvement of 10.98\% on F1. The path-based prompts can help to acquire patient-specific knowledge based on the graph, meanwhile assisting the inference. 

\paragraph{Error analysis.} We further investigate the causes of recommendation errors produced by CoT prompting. 30 inaccurate cases (with F1 $<$ 0.2) are randomly selected for analysis. We pre-define 3 types of recommendation errors, and three human expert evaluator are hired to identify the error types. As shown in Figure \ref{figure:error}, most errors derive from the overlooking of medical factors mentioned in dialogue history. The lack of medication knowledge occupies 30\% of the cases, leading to ineffective recommendations. Besides, LLMs sometimes excessively focus on non-essential medical information such as the past medical history of the patient, which results in useless recommendations likewise.

\paragraph{Case study.} Figure \ref{figure:cases} demonstrates the recommended results of CoT prompting and GAP on DialMed. The CoT prompt aims to analyze the disease, symptom of the patient and then provide corresponding medication recommendation. However, it generates an inaccurate treatment of \textit{dizziness} in the first case (\textit{Bicyclol} is used for anti-hepatitis virus replication), which leads to a non-factual recommendation. GAP concentrates on the main symptom \textit{diarrhea} and recommends accurate medications with the assistance of neighborhood prompts. In the second case, the CoT prompt results in a sub-optimal recommendation because \textit{Loratadine} should be used cautiously for pregnant women. In GAP, the neighborhood prompt provides several candidates for relieving \textit{itching}, and the path-based prompts elicited from KG and LLM further explained the rationale of selecting candidate medications, which jointly lead to the correct answer.


\section{Conclusion and Future Work}
We propose a Graph-Assisted Prompts (GAP) framework for dialogue-based medication recommendation. GAP maintains a patient-centric dialogue graph through information retrieval techniques and integrates it with KGs to acquire more information. GAP constructs two kinds of prompts to obtain patient-specific knowledge for response generation.
Experimental results present that our proposed framework outperforms strong baselines on different tasks and benchmark datasets. 
More importantly, this work further exploits the potential of utilizing LLMs as reliable and useful doctors.


%
%
%
\bibliographystyle{splncs04} 
\bibliography{latex/gap}

\clearpage

\appendix
\section{Appendix}
\subsection{Details for Patient-Centric Graph}
The patient-centric graph contains a patient node, patient characteristics, and medical information. For patient characteristics, we extract gender, age, blood pressure, and pregnant state if mentioned in the dialogue. For medical information, we mainly focus on the extraction of diseases, symptoms, and medications. 
The disease and symptom have the following slots and values: 
\begin{itemize}
    \item \textbf{state}: ["patient claims positive", "patient claims negative", "doctor claims positive", "doctor claims negative", "unknown"]
    \item \textbf{past medical history}:  ["yes", "no", "unknown"]
\end{itemize}
The medication has the following slots and values: 
\begin{itemize}
    \item \textbf{whether taken}: ["yes", "no", "unknown"]
    \item \textbf{whether effective}:  ["yes", "no", "unknown"]
    \item \textbf{whether doctor recommend}:  ["yes", "no", "unknown"]
\end{itemize}

We further extract relevant information about the medical concepts as memory which does not belong to the above fields.

\begin{figure}[]
    \centering
    \includegraphics[width=0.5\linewidth]{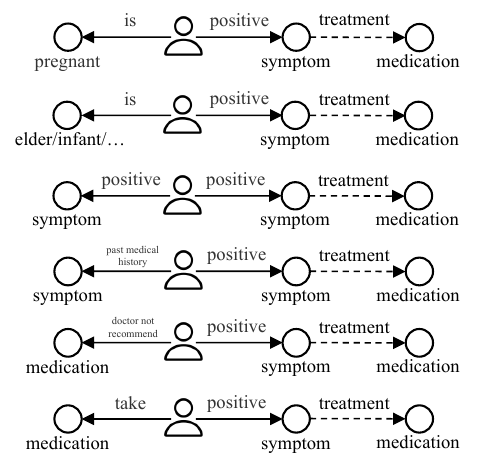}
    \caption{Prominent schema used for constructing path-based prompts.}
    \label{figure:schema}
\end{figure}

\subsection{Details for Medical Path Schema}
\label{appendix.schema}

The path schema used for generating path-based prompts is shown in Figure \ref{figure:schema}. Our main idea is to acquire or exclude possible medications based on the  
medical condition of the patient. The subsequently constructed path-based prompts which contain context-relevant knowledge can assist the medication recommendation. We can also change the target from acquiring the treatments to suitable food, medical tests, etc., therefore adapting the schema to fit the diagnostic interviewing scenario.

\subsection{Details for LLMs}
We use ChatGPT-3.5-Turbo (version 1106) for the experiments. The prompts we used in the experiments are shown in Table \ref{table:prompts_mie}-\ref{table:prompts_gap}.

\subsection{Details for Error Analysis}
\label{appendix.error_analysis}

We pre-define 3 types of recommendation errors on DialMed. Following are the definitions:

\begin{itemize}
    \item \textbf{Overlooking Medical Factors}: the LLM provides a recommendation that fits the symptoms but conflicts with some medical information mentioned in the dialogue history. E.g., Loratadine for relieving itching during pregnancy.
    \item \textbf{Lack of Knowledge about Medications}: the LLM provides wrong recommendations for the patient's symptoms or disease. E.g., Bicyclol for treating dizziness.
    \item \textbf{Overlooking the Main Symptoms}: the patient inquires about medications about one symptom, but the LLM focuses on another mentioned in the dialogue, such as the past medical history of the patient.
\end{itemize}


\begin{table*}[h]
\caption{Prompts used for medical concepts (disease/symptom/medication) extraction.}
\centering
\renewcommand{\arraystretch}{1.6}
\resizebox{\linewidth}{!}{
\begin{tabular}{p{20cm}}
\hline\hline
\textbf{Prompts}              \\ \hline

You are a named entity recognition (NER) annotator in the medical domain. Given a piece of medical dialogue context, you are required to return the existing disease/symptom/medication entities in list format as follows: \\
Output format: ["entity1", "entity2", ...] \\
The definition of disease/symptom/medication: [Definition]\\

[Demonstrations] \\ \\

Input context:\\

[Context] \\
Output result:
\\ \hline\hline
\end{tabular}
}
\label{table:prompts_mie}
\end{table*}

\begin{table*}[h]
\caption{Prompts used for judging the states of given concept.}
\centering
\renewcommand{\arraystretch}{1.6}
\resizebox{\linewidth}{!}{
\begin{tabular}{p{20cm}}
\hline\hline
\textbf{Prompts}              \\ \hline

\#You are an experienced doctor. Identify the states of the given disease/symptom based on the medical dialogue context, and return the result in JSON format. \\
\#\#main-state: candidate types come from ["patient-positive", "patient-negative", "doctor-positive", "doctor-negative", "unknown"] \\

[Descriptions of main-state types] \\
\#\#past medical history: candidate types come from ["yes", "no"] \\

[Descriptions of past medical history types] \\
\#\#other relevant information: other information about the given disease/symptom mentioned in the dialogue, such as duration and body parts. store the information in list format \\

[Demonstrations] \\

\\
\#\#\#Input context:\\

[Context] \\
\#\#\#The disease/symptom:\\

[Disease/Symptom] \\
\#\#\#Output:\\\hline\hline
\end{tabular}
}
\label{table:prompts_sv_pair}
\end{table*}

\begin{table*}[h]
\caption{Acquire knowledge from LLMs to generate path-based prompts.}
\centering
\renewcommand{\arraystretch}{1.6}
\resizebox{\linewidth}{!}{
\begin{tabular}{p{20cm}}
\hline\hline
\textbf{Prompts}              \\ \hline

You are an experienced doctor. Please provide some effective suggestions for the following medical questions, within 50 words. \\

\\
Question: [Question]\\
Suggestions: 
\\ \hline\hline
\end{tabular}
}
\label{table:prompts_pp_llm}
\end{table*}

\begin{table*}[h]
\caption{Chain-of-Thoughts (CoT) prompt used in medication recommendation.}
\centering
\renewcommand{\arraystretch}{1.6}
\resizebox{\linewidth}{!}{
\begin{tabular}{p{20cm}}
\hline\hline
\textbf{Prompts}              \\ \hline

You are an experienced doctor. Given a piece of medical dialogue context, you are required to recommend the medication based on patient's disease and symptom. The diseases are from Respirator, Gastroenterology, Dermatology, including: [Disease]. Candidate medications are from [Medication]. You need to think step by step and generate your thoughts. Demonstrations are as follow: \\

[Demonstrations] \\

\\
Input context:\\

[Context] \\
Now generate your thoughts and answers, please make sure the answers are from the candidate medication list:
\\ \hline\hline
\end{tabular}
}
\label{table:prompts_cot}
\end{table*}

\begin{table*}[h]
\caption{The prompt of GAP used in medication recommendation.}
\centering
\renewcommand{\arraystretch}{1.6}
\resizebox{\linewidth}{!}{
\begin{tabular}{p{20cm}}
\hline\hline
\textbf{Prompts}              \\ \hline

You are an experienced doctor. Given a piece of medical dialogue context, you are required to recommend the medication based on patient's disease and symptom. The diseases are from Respirator, Gastroenterology, Dermatology, including: [Disease]. Candidate medications are from [Medication]. The patient-centric graph is a summary of the dialogue. The neighborhood prompts and path-based prompts can be viewed as relevant knowledge. You need to think step by step and generate your thoughts based on the context. Demonstrations are as follow: \\

[Demonstrations] \\

\\
Input context:\\

[Context] \\
Patient-centric graph:\\

[Graph] \\
Neighborhood Prompts:\\

[NP] \\
Path-based Prompts:\\

[PP] \\
Now generate your thoughts and answers, please make sure the answers are from the candidate medication list:
\\ \hline\hline
\end{tabular}
}
\label{table:prompts_gap}
\end{table*}

\end{document}